\documentclass[runningheads]{llncs}

\usepackage{eccv}

\usepackage{eccvabbrv}

\usepackage{graphicx}
\usepackage{booktabs}

\usepackage[accsupp]{axessibility}  

\usepackage[pagebackref,breaklinks,colorlinks,citecolor=eccvblue]{hyperref}
\usepackage{orcidlink}

\usepackage{subcaption}
\usepackage{multirow}
\usepackage{bm}
\usepackage{amsmath}

\begin{document}

\title{SENet: A Spectral Filtering Approach to Represent Exemplars for Few-shot Learning} 

\titlerunning{SENet: a spectral filtering approach to represent exemplars for few-shot learning}

\author{Tao Zhang\inst{1}\orcidlink{0000-0002-0281-9234} \and
	Wu Huang\inst{2}\orcidlink{0000-0002-2525-6454}}
\authorrunning{Tao Zhang et al.}
%
\institute{Chengdu Techman Software Co., Ltd., Chengdu,
	Sichuan, China
	\email{ztuestc@outlook.com}\\ \and
	Sichuan University, Chengdu, Sichuan, China\\
	\email{huangwu@scu.edu.cn}}

\authorrunning{Tao Zhang et al.}

\maketitle

\begin{abstract}
Prototype is widely used to represent internal structure of category for few-shot learning, which was proposed as a simple inductive bias to address the issue of overfitting. However, since prototype representation is normally averaged from individual samples, it can appropriately to represent some classes but with underfitting to represent some others that can be batter represented by exemplars. To address this problem, in this work, we propose Shrinkage Exemplar Networks (SENet) for few-shot classification. In SENet, categories are represented by the embedding of samples that shrink towards their mean via spectral filtering. Furthermore, a shrinkage exemplar loss is proposed to replace the widely used cross entropy loss for capturing the information of individual shrinkage samples. Several experiments were conducted on miniImageNet, tiered-ImageNet and CIFAR-FS datasets. The experimental results demonstrate the effectiveness of our proposed method.
\end{abstract}

\section{Introduction}
\label{sec:intro}

A key aspect of human intelligence is the ability to learn new conceptual knowledge from one or few samples through flexible representations of categories. However, deep learning models require large amounts of data to learn better category representations and improve generalization. With a small sample size, or especially with the few-shot settings \cite{Feifei2006}, deep learning model performance falls far short of human intelligence performance. One way to solve this problem is to make the model "learning to learn", and thus the meta-learning paradigm is proposed to achieve this goal. However, it is still difficult for most models to learn a good enough representation to achieve bias-variance trade-off due to the lack of sufficient information.

A debate in cognitive psychology is whether categories are represented as individual samples or abstracted  prototypes \cite{nosofsky2011generalized}. Two mainstream models, the exemplar model and the prototype model, have therefore been proposed for answering this question. The exemplar model, which represents categories by individual samples, is better to represent the categories when prototypes are absent, e. g., the class of poker of the same suit. Motivated by this, we suppose that the dilemma of whether categories are represented by prototypes or examples exists in supervised few-shot
\begin{figure}[h]
	\centering
	{\includegraphics[width=0.9\textwidth]{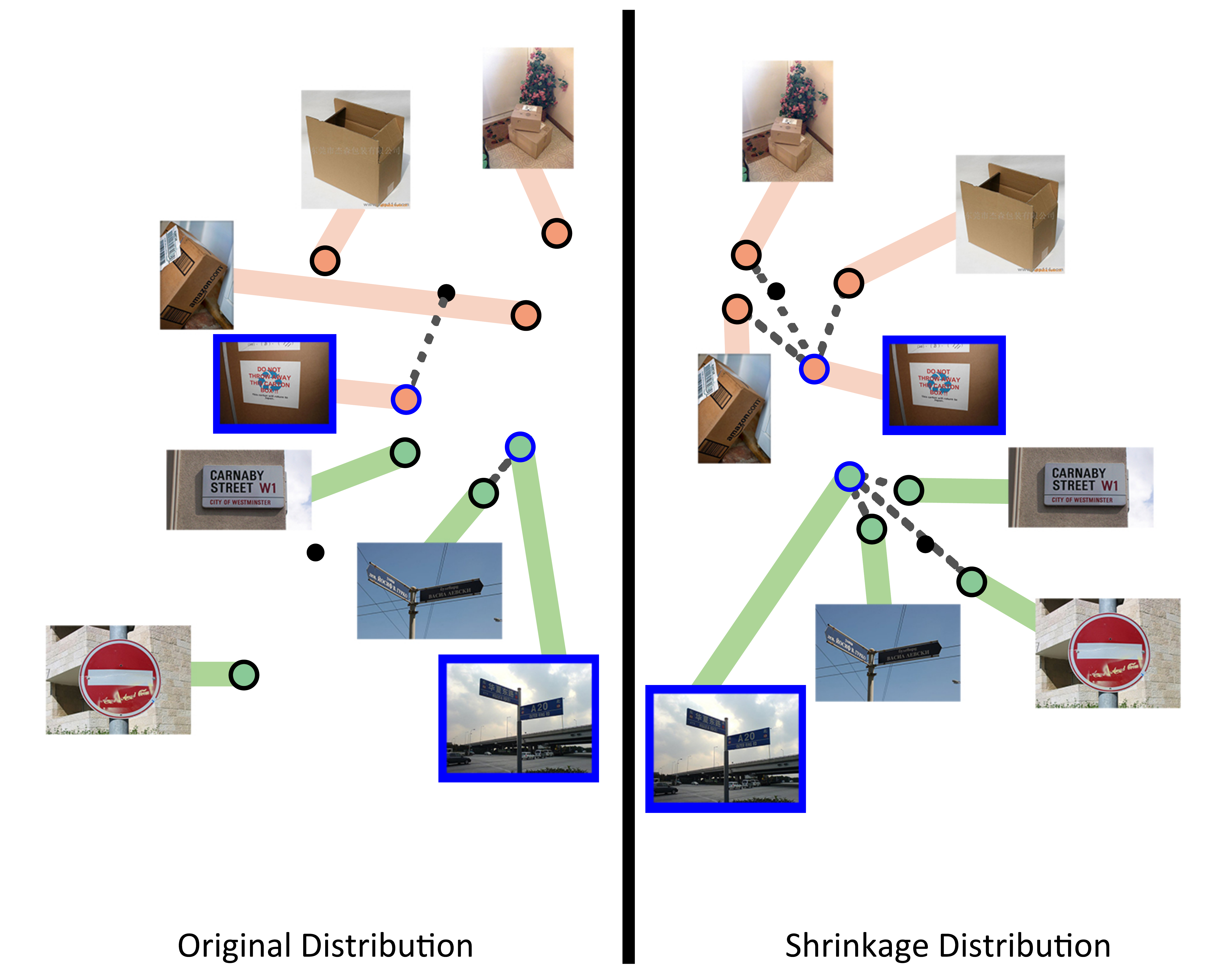}}
	\caption[]{Comparison of original distribution (left) to shrinkage distribution (right) for a 3-shot task in 2-dimension. Two categories including the cardboard box category (red) and the guide-board category (green) are considered, and the queries are annotated with blue box. In the original distribution,  the query belonging to cardboard box category should be predicted with the prototype model and that belonging to the guide-board category should be predicted with the exemplar model. In the shrinkage distribution, all the samples belonging to the same class shrink properly toward their mean. This case allow us to make a prediction uniformly via the similarities between queries and the shrinkage samples.}
	\label{fig:filter1}
\end{figure}
learning, meaning that using only prototypes or exemplars to represent all categories may not be a perfect strategy (See the left part of \cref{fig:filter1}). In this work, we propose the Shrinkage Exemplar Networks (SENet) from a perspective of filtering for few-shot learning. In this framework, categories with support set are represented by the embedding of shrinkage support examples, which comes from that the actual support samples shrink properly toward their mean (See the right part of \cref{fig:filter1}). These category representations of shrinkage support examples may perform better than those of actual support samples and prototypes thanks to Stein’s phenomenon \cite{efron1977stein}. The ``proper'' shrinkage means that, we need a hyper-parameter to adjust the degree of shrinkage, and thus exemplar representation and prototype representation are the two extreme cases of filtering. The shrinkage method draws on the relevant work of the shrinkage estimators, where examples shrink to different degrees on different components of principal component analysis (PCA).Furthermore, we uses the exemplar loss as the objective that is considered less than cross-entropy loss in few-shot learning. Our main contributions are three-fold:

1. From the perspective of filtering, we proposed a simple and novel model, the SENet, for few-shot learning, aiming to obtain the bias-variance trade-off when the sample size is insufficient.

2. We demonstrate that the proposed SENet outperforms the prototype model and the example model using different degrees of filtering on different benchmarks.

3. We propose to use shrinkage exemplar loss to replace the widely used cross entropy loss for few-shot learning, and prove that under the proposed loss form, the prototype-based model and the exemplar-based model are two different extreme filtering situations.

\section{Related Work}
\subsubsection{Few-shot Learning using Metric-based Model.} Meta-learning framework is widely used in few-shot learning tasks \cite{vinyals2016matching, snell2017prototypical,sung2018learning,finn2017model,rusu2018meta,simon2020adaptive,ravi2016optimization,oreshkin2018tadam,chen2020variational}. The metric-based meta-learning, as one of mainstream meta-learning frameworks, learn the similarities between the samples \cite{vinyals2016matching, snell2017prototypical,sung2018learning, simon2020adaptive,oreshkin2018tadam,chen2020variational}. A popular metric-based meta-learning is Matching Networks, where episodes is proposed for the purpose of matching between test and train conditions, and the labels of query samples are predicted based on labeled support examples using an attention mechanism \cite{vinyals2016matching}. For this approach, the original information of all support samples is retained and used for prediction, which makes the model prone to overfitting. To address this issue, Prototypical Networks and a family of its extension approaches was proposed, where prototype as a kind of inductive bias was introduced in few-shot learning to represent categories, and the similarities between a query sample and prototype representations are measured for classification, \cite{sung2018learning,chen2020variational, fort2017gaussian,pahde2021multimodal,liu2020prototype}. In addition, the similarities between query samples and prototypes (sample mean or sample sum) are given priority consideration, and thus more diverse classifiers have been developed \cite{sung2018learning,chen2020variational,zhang2022kernel,simon2020adaptive}. The prototype representation, however, may lead to underfitting in some cases. To address this issue, Infinite Mixture Prototypes (IMP) was proposed that represents categories as multiple clusters \cite{allen2019infinite}. Our work is similar to IMP in that it is all about figuring out how to properly balance examples and prototypes, but the difference is that we address this issue from a filtering perspective.

\subsubsection{Shrinkage Estimators.} Some previous work analyzed how to estimate the mean of a finite sample in reproducing kernel Hilbert space (RKHS)\cite{muandet2014kernel, muandet2017kernel, muandet2016kernel, muandet2014kernel}. They found that the kernel mean estimation can be improved thanks to the existence of Stein’s phenomenon, which implies that a shrinkage mean estimator, e.g., the James–Stein estimator, may achieves smaller risk than an empirical mean estimator \cite{efron1977stein, muandet2014kernel, stein1981estimation}. Based on this analysis,  a family of kernel mean shrinkage estimators, e. g., the spectral kernel mean shrinkage estimator, was designed to further improve the estimation accuracy using spectral filtering \cite{muandet2014kernel}. For simplicity, our work excludes the kernel method in estimating the example in embedding space using shrinkage estimator.

\subsubsection{Loss Objective.}  The design of loss is shown by many work to be crucial for improving performance of deep learning models. Among these losses, the cross-entropy loss is used very frequently in classification problems especially in few-shot classification. However, recent work pointed out that cross entropy loss has inadequate robustness to noise labels\cite{zhang2018generalized} and adversarial samples \cite{pang2019rethinking, feng2021can}. In deep metric learning, the loss functions, e. g.,  \cite{sohn2016improved, khosla2020supervised, chopra2005learning}, are usually used to draw samples with the same labels closer and push different samples farther.The proposed loss in our work is similar to some previous work that is shown in exemplar form, e. g., the supervised contrastive losses \cite{khosla2020supervised}. For these losses, the labeled samples as well as the augmentation-based samples are taken into account on the basis of self-supervised contrastive loss . In addition, the construction form of supervised contrastive losses is found to be critical, i.e., the position of $\log$ function in the loss can dramatically impacts the classification accuracy. 

\section{Methodology}
\subsection{Preliminary}

\subsubsection{Prototype-based predictors.} The prototype-based models focus on how to estimate abstracted prototypes to represent categories. For this approach, the possibility of a query sample belonging to a class depends on the similarity between a query sample and a prototype, i. e.

\begin{equation}\label{eq:Proto}
p_\theta(y'_l=c|\bm{q}_l)=\displaystyle \frac{\exp\left(- d\left(f_\theta(\bm{q}_l),\bm{\mu}_c\right)\right)}{\sum_{c'}\exp\left(- d\left(f_\theta(\bm{q}_l),\bm{\mu}_{c'}\right)\right)},
\end{equation}

where $\bm{\mu}_c$ the prototype representation of the $c$ class and $f_\theta(\bm{q}_l)$ the embedding of the $l$th query sample $\bm{q}_l$ with the label $y'_l$ after a mapping of network $f_\theta$ with the parameter set $\theta$. In addition, $d$ is the similarity distance that can be defined as, e. g.,  Euclidean distance and cosine distance. It is worth mentioning that the prototype representation $\bm{\mu}_c$ can not only be calculated straightforwardly by averaging the support samples in a class that $\bm{\mu}_{c} = \frac{1}{|\textbf{S}_{c}|} \sum_{\{\textbf{s}_{i},y_i\}\in \textbf{S}_{c}} f_\theta\left(\textbf{s}_{i}\right)$ where $\textbf{s}_{i}$ the $i$th support sample with the label $y_i$ and $\textbf{S}_{c}$ the support set labeled with class $c$ \cite{snell2017prototypical}, but also be calculated after rectification \cite{liu2020prototype} or adding cross-modal information \cite{xing2019adaptive}.

\subsubsection{Prototype-extended predictors.} In addition, there are some approaches that utilizes sample mean (sample sum) as prototypes, but they focus on how to measure the similarity, e. g., \cite{sung2018learning,oreshkin2018tadam,zhang2022kernel,simon2020adaptive}. Therefore, we call them as prototype-extended approaches. For example, Deep Subspace Networks projects the similarity distance between prototype and query into a subspace for prediction \cite{simon2020adaptive}. Relation Network sums over the embedding of support samples in a class (prototype) and learns the similarity between prototype and query samples \cite{sung2018learning}. Deep Spectral Filtering Networks applies the kernel shrinkage estimation for similarity measure between prototype and query samples \cite{zhang2022kernel}. In addition, TADAM learn a scale metric for better similarity measurement based on the Prototypical Networks \cite{oreshkin2018tadam}.

\subsubsection{Exemplar-based predictors.} In \cref{eq:Proto}, the prototype representation $\bm{\mu}_c$ captures the average features of the samples belonging to the class $c$ in an embedded space, but discards their individual features. Compared with the prototype model, the exemplar model can capture the original information of the samples, and the possibility of a query sample $\bm{q}_l$ belonging to a class $c$ is given as

\begin{equation}\label{eq:Exemplar1}
p(y'_l=c|\bm{q}_l)= \sum_{i:y_i=c}p_{li},
\end{equation}

with

\begin{equation}\label{eq:Exemplar2}
p_{li} = \displaystyle \frac{\exp\left(- d\left(f_\theta(\bm{q}_l),f_\theta(\textbf{s}_{i})\right)\right)}{\sum_{j}\exp\left(- d\left(f_\theta(\bm{q}_l),f_\theta(\textbf{s}_{j})\right)\right)}.
\end{equation}
A typical exemplar-based model is Matching Networks \cite{vinyals2016matching}. In this model, the exemplar representation is presented as a loss target based on an attention mechanism, and the similarity is measured using cosine distance.

\subsection{Shrinkage Exemplar}
Our proposed method is based on the example model in \cref{eq:Exemplar1} to make a shrinkage estimation for the support samples of each class. First, let $\textbf{S}_{c} = \{\bm{s}_{1,c},...,\bm{s}_{K,c}\}$ with $\bm{s}_{i,c}\in \mathbb{R}^{d\times 1}(i=1,...,K)$  denotes the support samples belonging to class $c$ and $\textbf{Q} = \{\bm{q}_{1},...,\bm{q}_{M}\}$ with $\textbf{q}_{j}\in \mathbb{R}^{d\times 1}(j=1,...,M)$ denotes the query set. Consider the exemplar model where the similarity measure between $\bm{q}_l$ and $\bm{s}_{i,c}$ is filtered by a matrix $\bm{M}_{\lambda,c}$, and thus the probability of $\bm{q}_l$ sharing the same label $c$ with $\bm{s}_{i,c}$ after spectral filtering can be obtained, i. e., 

\begin{equation}\label{eq:SExemplar}
p_{li}=\displaystyle \frac{\exp\left(- \alpha d_\lambda\left(f_\theta(\bm{q}_l), f_\theta(\textbf{s}_{i,c}) \right)\right)}{\sum_{c'}\sum_{j}\exp\left(- \alpha d_\lambda\left(f_\theta(\bm{q}_l), f_\theta(\textbf{s}_{i,c})\right)\right)},
\end{equation}

Where $d_\lambda(f_\theta(\bm{q}_l), f_\theta(\textbf{s}_{i,c}))$ with the shrinkage coefficient $\lambda$ ($0 \leq \lambda <\infty$) is the shrinkage distance. Here we define two forms of shrinkage distance in \cref{eq:SExemplar}:

\begin{equation}\label{eq:d1}
d_\lambda^{(s_1)}\left(f_\theta(\bm{q}_l), f_\theta(\textbf{s}_{i,c})\right)=\displaystyle ||\bm{M}_{\lambda, c}\left(f_\theta(\bm{q}_l) - f_\theta(\textbf{s}_{i,c}) \right)||^2,
\end{equation}

\begin{equation}\label{eq:d2}
\begin{split}
d_\lambda^{(s_2)}(f_\theta(\bm{q}_l), f_\theta(\textbf{s}_{i,c}))=\displaystyle ||\left(f_\theta(\bm{q}_l) - \bm{\mu}_{c} - \bm{M}_{\lambda, c} \left(f_\theta(\textbf{s}_{i,c}) - \bm{\mu}_{c}\right) \right)||^2,
\end{split}
\end{equation}

with

\begin{equation}\label{eq:filter}
\bm{M}_{\lambda,c} =  \bm{W}_cg(\bm{\Gamma}_c,\lambda)\bm{W}_c^T.
\end{equation}

In \cref{eq:d1} and \cref{eq:d2}, $\bm{\mu}_{c} = \frac{1}{|\textbf{S}_{c}|} \sum_{\{\textbf{s}_{i},y_i\}\in \textbf{S}_{c}} f_\theta\left(\textbf{s}_{i}\right)$. In \cref{eq:filter}, $\bm{\Gamma}_c= {\rm{diag}}(\gamma_{1,c},...,\gamma_{d,c})$ and $\bm{W}_c = \{\bm{w}_{1,c},...,\bm{w}_{d,c}\}$ where $\gamma_{i,c}$  and $\bm{w}_{k,c}$ are respectively the $i$th eigenvalue and eigenvector of the empirical covariance matrix of support samples belonging to class $c$: $\bm{M}_{\lambda,c} = \sum_{i} (f_\theta(\textbf{s}_{i,c}) - \bm{\mu}_c) \otimes (f_\theta(\textbf{s}_{i,c}) - \bm{\mu}_c)$ \footnote[1]{The eigenvectors $\bm{W}_c$ are unitized.}. For the filter $g$, the diagonal entries of $\bm{\Gamma}_c$ are mapped specifically in the following way \footnote[2]{The filter $g$ in \cref{eq:SS} is designed referring to Tikhonov regularization, and in addition to this, there is also a family of filters for reference. See \cite{muandet2014kernel}.}:

\begin{equation}\label{eq:SS}
g_e(\gamma_{n,c},\lambda) = \left\{
\begin{aligned}
\frac{\gamma_{n,c}}{\lambda + \gamma_{n,c}} \quad \gamma_{n,c} \neq 0\\
1 \quad \quad \quad \gamma_{n,c} = 0\\
\end{aligned}
\right.
\end{equation}

In \cref{eq:SS}, as $\lambda = 0$, no filtering is performed, and as $\lambda$ increases, the filtering degree becomes deeper. In the extreme, as $\lambda \to \infty$, all support samples merge into one sample. In \cref{eq:d1}, $d^{(s_1)}$ implies the filtering of both support samples and query samples, which can be seen as a process of filtering out noise or irrelevant components. In \cref{eq:d2}, $d^{(s_2)}$ implies that only support samples is filtered, which can be seen as the process of extracting common representations of categories. Based on these two kinds od filters respectively, our loss objective can be expressed as the exemplar form:

\begin{equation}\label{eq:SExemplarL1}
\mathcal{L} =-\displaystyle \frac{1}{M}\sum_{l=0}^M \log \sum_{i:y_i=y'_l}   p_{li}.
\end{equation}

The loss in \cref{eq:SExemplarL1} implies that we need to maximize the mean logarithm of the probability of correctly classifying all queries.

\subsubsection{Temperature.} Here we analyze the role of temperature in our model. The impact of temperature on prototype-based predictors, e. g., TADAM \cite{oreshkin2018tadam}, is analyzed in detail. It mathematically proves that learning a suitable temperature is equivalent to hard sample mining based on the embedding distance between query samples and prototypes. As well as cross entropy loss, the temperature in supervised contrastive losses also plays a role of learner for hard sample mining but based on the embedding distance between samples \cite{khosla2020supervised}. For facilitate analysis, the cases of $\lambda = 0$ and $\lambda \to \infty$ in the derivative of loss in our proposed model is derived :

\begin{equation}\label{eq:p2}
\begin{split}
\lim_{\alpha \to \infty}\frac{1}{\alpha}\frac{\partial\mathcal{L}}{\partial\theta} = \displaystyle \frac{1}{M}\sum_{l=0}^M \left( \frac{\partial d_\lambda\left(f_\theta(\bm{q}_l), f_\theta(\textbf{s}_{i,c}^*)\right)}{\partial \theta} -{ \frac{\partial d_\lambda\left(f_\theta(\bm{q}_l), f_\theta(\textbf{s}_{j,c'}^*)\right)}{\partial \theta}}\right)
\end{split}
\end{equation}

\begin{equation}\label{eq:p1}
\begin{split}
\lim_{\alpha \to 0}\frac{1}{\alpha}\frac{\partial\mathcal{L}}{\partial\theta} = \displaystyle \frac{1}{M}\sum_{l=0}^M \left(         \frac{1}{K}\sum_{i:y_i=y'_l} \frac{\partial d_\lambda\left(f_\theta(\bm{q}_l), f_\theta(\textbf{s}_{i,c})\right)}{\partial \theta} - \frac{1}{CK} {\sum_{c'}\sum_{j} \frac{\partial d_\lambda\left(f_\theta(\bm{q}_l), f_\theta(\textbf{s}_{j,c'})\right)}{\partial \theta}}\right)\\
= \displaystyle \frac{1}{M}\sum_{l=0}^M \left(         \frac{C-1}{CK}\sum_{i:y_i=y'_l} \frac{\partial d_\lambda\left(f_\theta(\bm{q}_l), f_\theta(\textbf{s}_{i,c})\right)}{\partial \theta} - \frac{1}{CK} {\sum_{c' \neq y'_l}\sum_{j} \frac{\partial d_\lambda\left(f_\theta(\bm{q}_l), f_\theta(\textbf{s}_{j,c'})\right)}{\partial \theta} }\right)\\
\end{split}
\end{equation}

In \cref{eq:p1}, we can see that as $\alpha = 0$, the distances between supports and a query with the same labels should decrease, and the distances between supports and a query with the different labels should increase. In \cref{eq:p2}, as $\alpha \to \infty$, if the distance between closest support $\textbf{s}_{i,c}^*$ and a query $\textbf{q}_{l}^*$ with the same label should decrease and that between closest support $\textbf{s}_{j,c'}^*$ and a query $\textbf{q}_{l}^*$ with different label should increase if $\textbf{s}_{i,c}^* \neq \textbf{s}_{j,c'}^*$. If $\textbf{s}_{i,c}^* = \textbf{s}_{j,c'}^*$, implying that the closest sample of a query sharing the same label, these is no error. These results are similar to those in TADAM, but the temperature as is learner for hard sample mining in our model is based on the distances between supports and queries rather than those between prototypes and queries.

\subsection{Connection to Exemplar-based and Prototype-extended Predictors}

Our proposed model in \cref{eq:SExemplar} does not filter the data as $\lambda=0$, and thus it is equivalent to an exemplar-based predictor; When $\lambda\to\infty$, our proposed model with $d_\lambda^{(s_1)}$  is reduced to the Deep Subspace Networks with no regularization \cite{simon2020adaptive} 

\begin{equation}\label{eq:SExem_1}
\begin{aligned}
\lim_{\lambda \to \infty}\mathcal{L}(\lambda) =\displaystyle - \sum_{l=0}^M \log \frac{\exp\left(- \alpha||\bm{B}_c\bm{B}_c^T\left(f_\theta(\bm{q}_l) - \bm{\mu}_c\right) ||^2\right)}{\sum_{c'}\exp\left(- \alpha ||\bm{B}_{c'}\bm{B}_{c'}^T\left(f_\theta(\bm{q}_l) - \bm{\mu}_{c'}\right)||^2\right)},
\end{aligned}
\end{equation}
where $\bm{B}$ is the truncation matrix, and  with $d_\lambda^{(s_2)}$ is reduced to ProtoNet \cite{snell2017prototypical}
\begin{equation}\label{eq:SExem_2}
\begin{aligned}
\lim_{\lambda \to \infty}\mathcal{L}(\lambda) =\displaystyle  -\sum_{l=0}^M \log \frac{\exp\left(- \alpha||\left(f_\theta(\bm{q}_l) - \bm{\mu}_c\right) ||^2\right)}{\sum_{c'}\exp\left(- \alpha ||\left(f_\theta(\bm{q}_l) - \bm{\mu}_{c'}\right)||^2\right)}.
\end{aligned}
\end{equation}

As $\lambda = 0$, the proposed SENet with $d^{(s_1)}_\lambda$ and $d^{(s_2)}_\lambda$ are equivalent to an exemplar-based predictor,

\begin{equation}\label{eq:Sproto1}
\begin{aligned}
\mathcal{L}(\lambda=0)= \displaystyle -\sum_{l=0}^M \log \sum_{i=0}^K \frac{\exp\left(- \alpha||f_\theta(\bm{q}_l)  -  f_\theta(\textbf{s}_{i,c})||^2\right)}{\sum_{c'}\sum_{j}\exp\left(- \alpha ||f_\theta(\bm{q}_l) -  f_\theta(\textbf{s}_{j,c'})||^2\right)},
\end{aligned}
\end{equation}

\begin{figure}[h]
	\centering
	\begin{subfigure}{0.3\textwidth}
	\includegraphics[width=\linewidth]{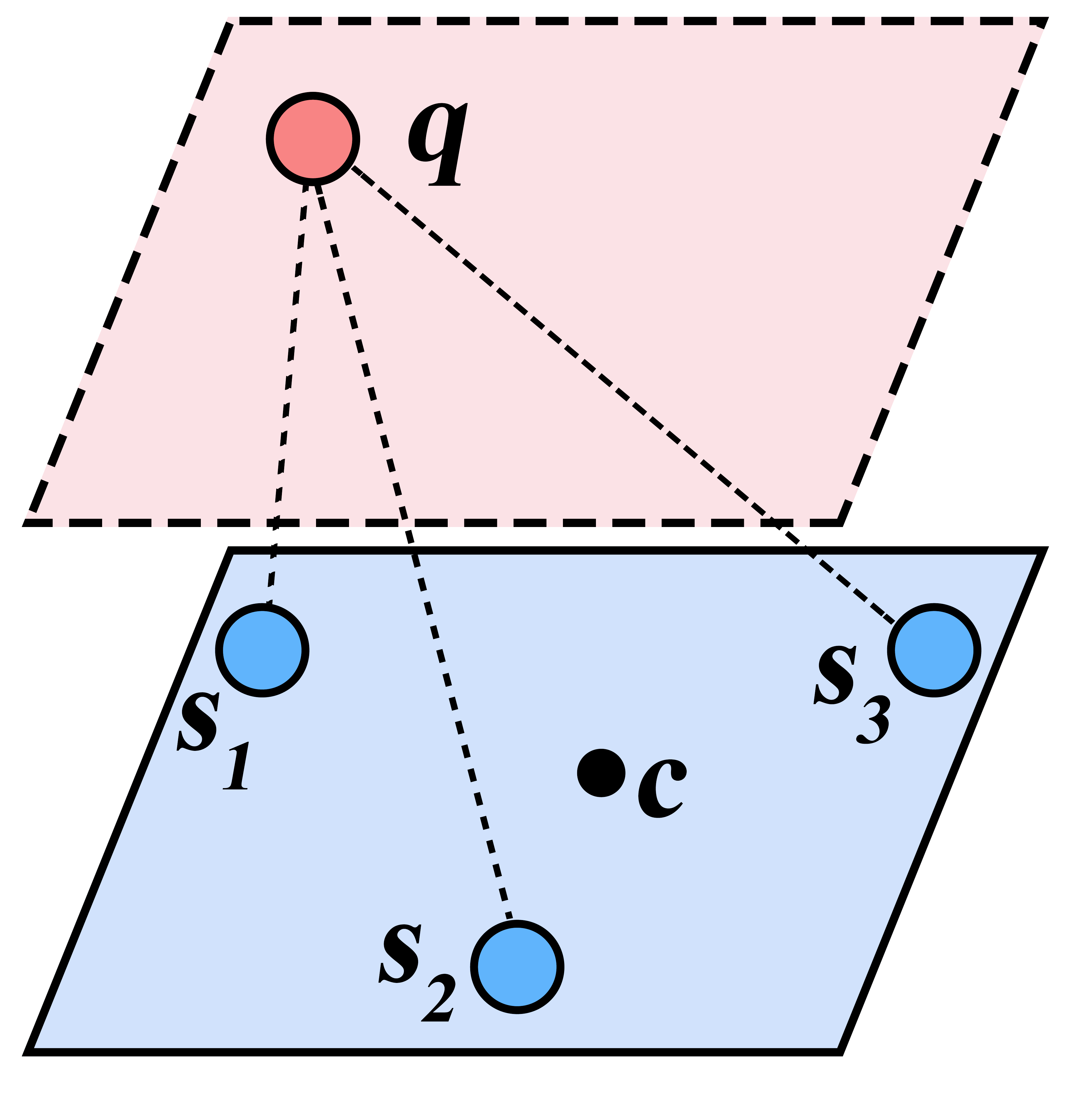}
	\caption[Exemplar]{}
\end{subfigure}
\begin{subfigure}{0.3\textwidth}
	\includegraphics[width=\linewidth]{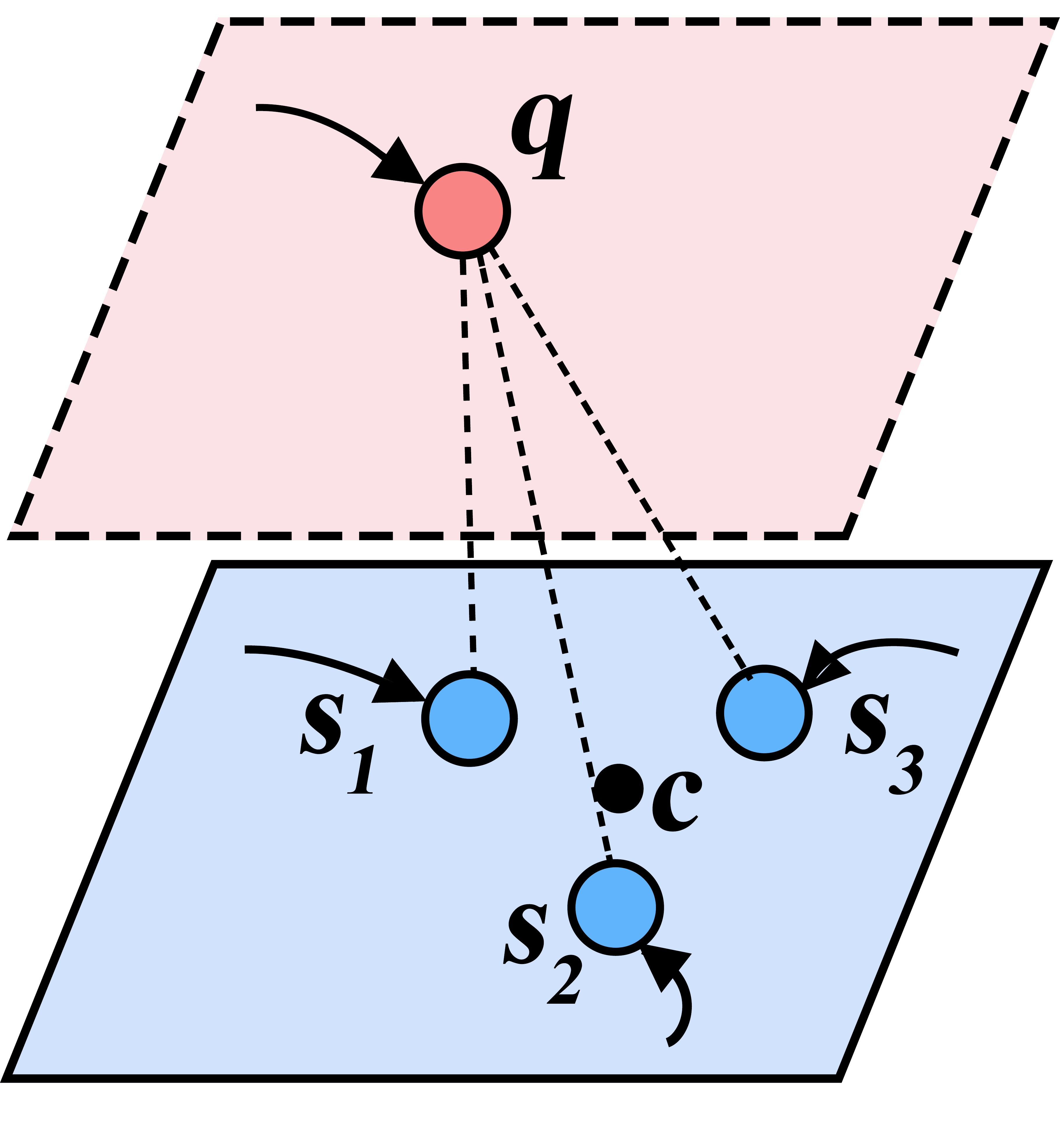}
	\caption[Filtering]{}
\end{subfigure}
\begin{subfigure}{0.3\textwidth}
	\includegraphics[width=\linewidth]{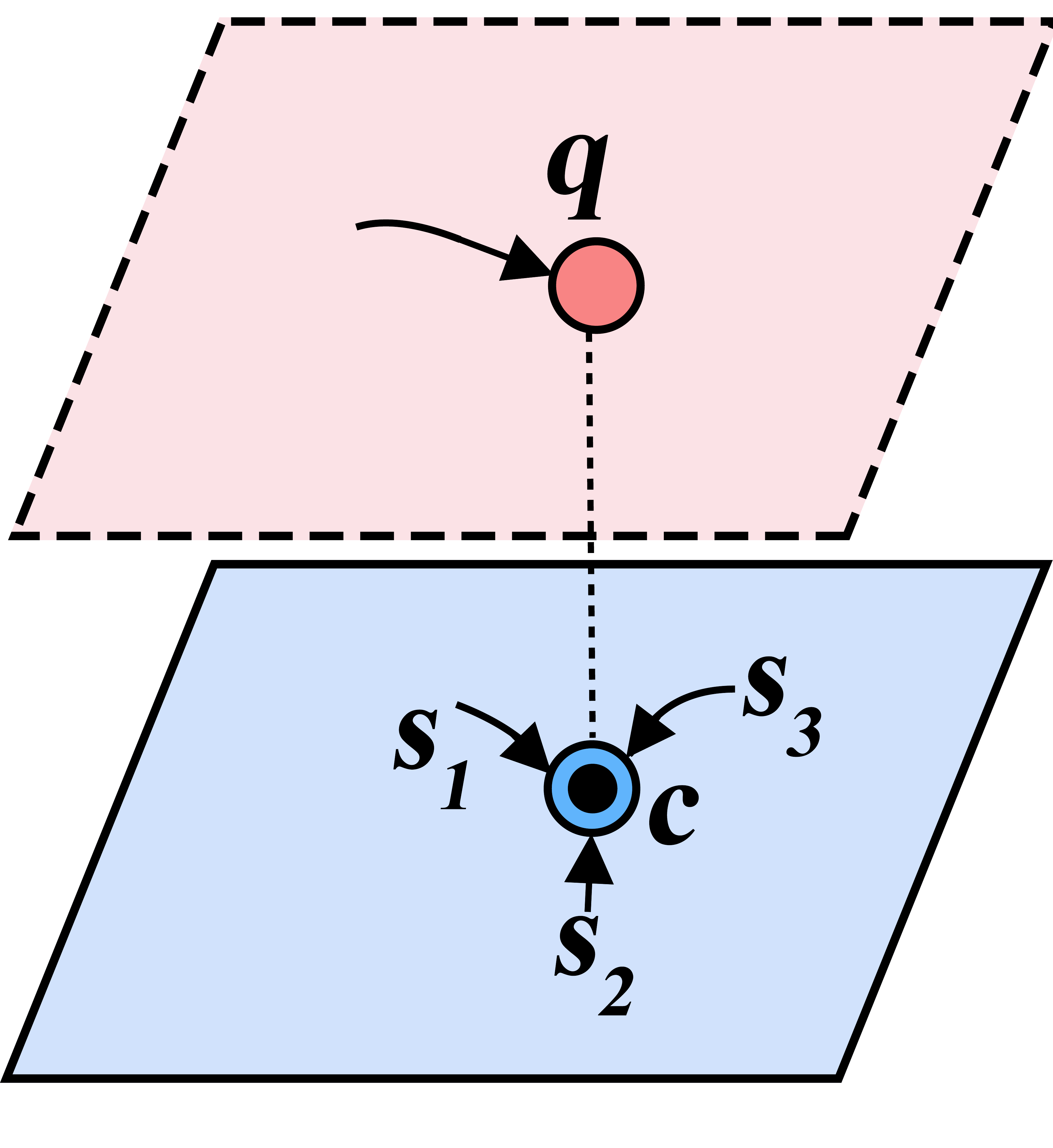}
	\caption[Prototype]{}
\end{subfigure}
	\caption[]{Comparison between three kinds of predictors in the case of 3-shot and 3 dimension. (a) Exemplar: no filtering on the support samples $s_1$, $s_2$, $s_3$ and the query samples $q$. (b) Filtering: $s_1$, $s_2$, $s_3$ and $q$ shrink towards the mean value of $s_1$, $s_2$, $s_3$ after spectral filtering. (c) Prototype: $s_1$, $s_2$, $s_3$ and the projection of $q$ on the subspace spanned by these support samples arrive the mean value after extreme filtering.}
	\label{fig:compare}
\end{figure}

The intuitive comparison between the three models is shown in  \cref{fig:compare}. Our model can filter data flexibly by setting $\lambda$ in the range of $0 \leq \lambda <\infty$, by which it may be possible to achieve the bias-variance trade-off in data distributions with different complexity.  One would naturally think that using multiple prototypes is also a flexible way to adjust inductive bias. However, the prototype number is severely limited by the sparsity of data and as a hyper-parameter can not be adjusted smoothly.

\section{Experiments}
\subsection{Experimental Setup}
\subsubsection{Datasets.} We use \emph{mini}ImageNet \cite{vinyals2016matching}, \emph{tiered}-ImageNet \cite{mishra2017simple} and CIFAR-FS \cite{bertinetto2018meta} as the datasets in our experiments. \emph{mini}ImageNet contains 84
$\times$ 84 images that come from ILSVRC-2012, which is divided into 100 classes and 600 images in each class. The \emph{tiered}-ImageNet also contains 84 $\times$ 84 images that come from ILSVRC-2012, which is divided into contains 608 classes that is more than those in \emph{mini}ImageNet. Each classes are divided into 34 high-level categories that contains 10 to 30 classes. The CIFAR-FS \cite{bertinetto2018meta} contains 32 $\times$ 32 images from CIFAR-100 \cite{krizhevsky2009learning} that are divided into 100 classes and each class contains 600 images. 

\subsubsection{Implementation details.} We choose 15-shot 10-query samples on \emph{mini}ImageNet dataset, \emph{tiered}-ImageNet dataset and CIFAR-FS dataset for 1-shot task and 5-shot task in training stage. Our model undergoes 80 training epochs, with each epoch comprising of 1000 sampled batches. During the testing phase, our model is evaluated with 1000 episodes. Since spectral filtering requires that the number of samples is more than one, we use flipping method to perform data augmentation on the original data for  1-shot tasks. Our code runs in the PyTorch machine learning package \cite{paszke2017automatic} throughout the entire phase. In addition, the backbone in our model is Resnet-12 \cite{he2016deep} for most cases, and 
the SGD optimizer \cite{bottou2010large} is applied for training the models. We set the learning rate initially as 0.1 then adjust it to 0.0025,0.00032, 0.00014 and 0.000052 at 12, 30, 45 and 57 epochs, respectively.  

\subsection{Comparison to State-of-the-art}
The comparison of the performance of proposed SENet to those of the state-of-the-art methods is shown in \cref{tab:mini}, \cref{tab:tiered} and \cref{tab:CIFAR} on \emph{mini}ImageNet, \emph{tiered}-ImageNet and CIFAR-FS datasets, respectively. Here, the setting of 8 episodes per batch is utilized on all the datasets. \cref{tab:mini} and \cref{tab:CIFAR} shows that for 5-shot tasks, the proposed SENet can be the state-of-the-art. In \cref{tab:tiered} for 5-shot tasks, the SENet performs the second best on \emph{tiered}-ImageNet datasets. For 1-shot tasks in \cref{tab:mini}, \cref{tab:tiered} and \cref{tab:CIFAR}, SENet has no obvious advantages compared with other state-of-the-art methods, and the reason would be that filters require capturing the interrelationships between samples, and one single support sample can not provide sufficient information for this filtering. 

\begin{table}[h]
	\centering
	\setlength\tabcolsep{2pt}
		\caption{Accuracy ($\%$) comparison with the state-of-the-art for 5-way tasks on $\textit{mini}$ImageNet with 95$\%$ confidence intervals. $\ddag$ denotes that validation set as well as training set is used for training. }
		\begin{tabular}{cccc}
			\toprule[1pt]
			
			\textbf{} 
			\multirow{2}{*}{\bf{Model}} & \multirow{2}{*}{\bf{Backbone}} &\multicolumn{2}{c}{\bf{\textit{mini}ImageNet}}\\
			\cmidrule(r){3-4}
			& &\bf{1-shot} &\bf{5-shot} \\
			\hline
			ProtoNet\cite{snell2017prototypical}           &ResNet-12         &$59.25_{\pm 0.64}$&$75.60_{\pm 0.48}$ \\
			SNAIL\cite{mishra2017simple}                       &ResNet-12         &$55.71_{\pm 0.99}$   &$68.88_{\pm 0.92}$   \\  
			TADAM(Oreshkin et al. 2018)                &ResNet-12         &$58.50_{\pm 0.30}$   &$76.70_{\pm 0.30}$   \\
			AdaResNet\cite{munkhdalai2018rapid}           &ResNet-12         &$56.88_{\pm 0.62}$   &$71.94_{\pm 0.57 }$  \\
			LEO$\ddag$\cite{rusu2018meta}                 &WRN-28-10         &$61.76_{\pm 0.08 }$   &$77.59_{\pm 0.12 }$  \\
			LwoF\cite{gidaris2018dynamic}                 &WRN-28-10         &$60.06_{\pm 0.14 }$   &$76.39_{\pm 0.11 }$  \\
			DSN\cite{simon2020adaptive}                   &ResNet-12         &$\bm{62.64_{\pm 0.66}}$   &$78.83_{\pm 0.45}$ \\  
			CTM\cite{li2019finding}                       &ResNet-18         &$62.05_{\pm 0.55}$   &$78.63_{\pm 0.06}$  \\
			Baseline\cite{chen2019closer}                 &ResNet-18         &$51.75_{\pm 0.80}$   &$74.27_{\pm 0.63}$  \\
			Baseline++\cite{chen2019closer}               &ResNet-18         &$51.87_{\pm 0.77}$   &$75.68_{\pm 0.63}$ \\
			Hyper ProtoNet\cite{khrulkov2020hyperbolic}   &ResNet-18         &$59.47_{\pm 0.20}$   &$76.84_{\pm 0.14}$  \\	
			MetaOptNet-RR\cite{lee2019meta}      &ResNet-12       &$61.41_{\pm 0.61} $&$77.88_{\pm 0.46}$ \\
			MetaOptNet-SVM\cite{lee2019meta}      &ResNet-12      &$\bm{62.64_{\pm 0.61}}$   &$78.63_{\pm 0.46}$\\
			\hline
			SENet($d^{(s_1)}_\lambda$ with optimal $\lambda$)     &ResNet-12  &$61.67_{\pm 0.65}$   &$\bm{80.04_{\pm 0.49}}$ \\
			SENet($d^{(s_2)}_\lambda$ with optimal $\lambda$)     &ResNet-12   &$61.07_{\pm 0.67} $   &$79.03_{\pm 0.52}$ \\
			\bottomrule[1pt]
			
		\end{tabular}\label{tab:mini}
\end{table}
\begin{table}[h]
	\centering
	\setlength\tabcolsep{2pt}
		\caption{Accuracy ($\%$) comparison with the state-of-the-art for 5-way tasks on $\textit{tiered}$-ImageNet with 95$\%$ confidence intervals. $\ddag$ denotes that validation set as well as training set is used for training. }
		\begin{tabular}{cccc}
			\toprule[1pt]
			
			\textbf{} 
			\multirow{2}{*}{\bf{Model}} & \multirow{2}{*}{\bf{Backbone}}	&\multicolumn{2}{c}{\bf{\textit{tiered}-ImageNet}}\\
			\cmidrule(r){3-4}
			& &\bf{1-shot} &\bf{5-shot} \\
			\hline
			ProtoNet\cite{snell2017prototypical}          &ResNet-12   &$61.74_{\pm 0.77 }$   &$80.00_{\pm 0.55}$  \\  
			LEO$\ddag$\cite{rusu2018meta}                 &WRN-28-10     &$\bm{66.33_{\pm 0.05 }}$   &$81.44_{\pm 0.09 }$  \\
			DSN\cite{simon2020adaptive}                   &ResNet-12     &$66.22_{\pm 0.75 }$   &$\bm{82.79_{\pm 0.48 }}$  \\  
			CTM\cite{li2019finding}                       &ResNet-18     &$64.78_{\pm 0.11}$   &$81.05_{\pm 0.52}$ \\	
			MetaOptNet-RR\cite{lee2019meta}      &ResNet-12      &$65.36_{\pm 0.71}$   &$81.34_{\pm 0.52}$ \\
			MetaOptNet-SVM\cite{lee2019meta}      &ResNet-12     &$65.99_{\pm 0.72}$   &$81.56_{\pm 0.53}$ \\
			\hline
			SENet($d^{(s_1)}_\lambda$ with optimal $\lambda$)    &ResNet-12      &$65.30_{\pm 0.71}$&$82.21_{\pm 0.69}$\\
			SENet($d^{(s_2)}_\lambda$ with optimal $\lambda$)    &ResNet-12     &$64.88_{\pm 0.72}$ &$81.20_{\pm 0.55}$ \\
			\bottomrule[1pt]
			
		\end{tabular}\label{tab:tiered}
\end{table}
\begin{table}[h]
	\centering
	\setlength\tabcolsep{2pt}
		\caption{Accuracy ($\%$) comparison with the state-of-the-art for 5-way tasks on CIFAR-FS with 95$\%$ confidence intervals. }
		\begin{tabular}{cccc}
			\toprule[1pt]
			
			\textbf{} 
			\multirow{2}{*}{\bf{Model}} & \multirow{2}{*}{\bf{Backbone}} 
			&\multicolumn{2}{c}{\bf{CIFAR-FS}}\\
			\cmidrule(r){3-4}
			& &\bf{1-shot} &\bf{5-shot} \\
			\hline
			ProtoNet\cite{snell2017prototypical}          &ResNet-12      &$72.2_{\pm 0.7 }$      &$83.5_{\pm 0.5 }$   \\
			DSN\cite{simon2020adaptive}                   &ResNet-12    &$72.3_{\pm 0.7 }$      &$85.1_{\pm 0.5 }$ \\  
			Baseline\cite{chen2019closer}                 &ResNet-18    &$65.51_{\pm 0.87}$&$82.85_{\pm 0.55}$\\
			Baseline++\cite{chen2019closer}               &ResNet-18    &$67.02_{\pm 0.90}$&$83.58_{\pm 0.54}$\\
			Hyper ProtoNet\cite{khrulkov2020hyperbolic}   &ResNet-18    &$64.02_{\pm 0.24}$&$82.53_{\pm 0.14}$\\	
			MetaOptNet-RR\cite{lee2019meta}      &ResNet-12     &$\bm{72.6_{\pm 0.7}}$      &$84.3_{\pm 0.5}$ \\
			MetaOptNet-SVM\cite{lee2019meta}      &ResNet-12     &$72.0_{\pm 0.7}$      &$84.2_{\pm 0.5}$ \\
			\hline
			SENet($d^{(s_1)}_\lambda$ with optimal $\lambda$)   &ResNet-12      &$71.63_{\pm 0.73}$&$\bm{86.23_{\pm 0.48}}$  \\
			SENet($d^{(s_2)}_\lambda$ with optimal $\lambda$)   &ResNet-12     &$70.59_{\pm 0.72}$&$85.64_{\pm 0.47}$  \\
			\bottomrule[1pt]
			
		\end{tabular}\label{tab:CIFAR}
\end{table}

\subsection{Ablation Study}
In this section, we experimentally compare the performances of the prototype model, the exemplar model and our proposed SENet to demonstrate the superiority of our model over the other two models. Since adjusting the shrinkage coefficient can make our model degenerate into the other two models, these three models will be obtained by our model under different shrinkage coefficients for the sake of fairness of comparison.

Firstly, the impact of $\lambda$ on the performance of SENet is analysed numerically. For example, we list the accuracies obtained by SENet with different $\lambda$ in \cref{tab:As}. On $\textit{mini}$ImageNet and $\textit{tiered}$-ImageNet, the number of episodes per batch is 4 and On CIFAR-FS, the number of episodes per batch is 8.  For the 5-shot tasks, SENet performs the best as $\lambda = 10^4$ on CIFAR-FS and $\lambda = 10^5$ on $\textit{mini}$ImageNet, $\textit{tiered}$-ImageNet. For the 1-shot tasks, SENet performs the best at the range of $1 \leq  \lambda \leq 1000$.  In addition, the tendency of accuracy for 1-shot tasks is not obvious, and for 5-shot tasks, accuracy shows an upward trend before reaching its maximum value. 

\begin{table}[h]
	\centering
	\scriptsize
		\caption{Accuracies ($\%$) with different shrinkage parameter $\lambda$ using ResNet-12 for 5-way tasks on $\textit{mini}$ImageNet,$\textit{tiered}$-ImageNet and CIFAR-FS with 95$\%$ confidence intervals.}
		\begin{tabular}{ccccccccc}
			\toprule[1pt]
			\textbf{} 
			\multirow{2}{*}{\bf{Type}}
			&\multirow{2}{*}{\bf{Shrinkage}} 
			&\multicolumn{2}{c}{\bf{\textit{mini}ImageNet}} 
			&\multicolumn{2}{c}{\bf{\textit{tiered}-ImageNet}} 
			&\multicolumn{2}{c}{\bf{CIFAR-FS}}\\
			
			\cmidrule(r){3-4} \cmidrule(r){5-6} \cmidrule(r){7-8}
			
			&& \bf{1-shot} &\bf{5-shot} &\bf{1-shot} &\bf{5-shot} &\bf{1-shot}  &\bf{5-shot}\\
			\hline
			&$\lambda=10^0$  &$61.56_{\pm 0.68}$ &$73.91_{\pm 0.56}$  &$\bm{63.27_{\pm 0.72}}$ &$75.48_{\pm 0.69}$ &$71.46_{\pm 0.74}$   &$82.79_{\pm 0.53} $   \\
			&$\lambda=10^1$   &$\bm{61.81_{\pm 0.68}}$   &$74.34_{\pm 0.55}$   &$62.29_{\pm 0.72}$   &$74.69_{\pm 0.60}$ &$71.65_{\pm 0.74} $   &$83.91_{\pm  0.51}$   \\
			&$\lambda=10^2$   &$61.49_{\pm 0.66}$   &$73.97_{\pm 0.54}$    &$61.61_{\pm 0.72}$   &$74.20_{\pm 0.59}$ &$\bm{72.06_{\pm 0.72}}$   &$83.85_{\pm  0.51}$   \\	
			$d^{(s_1)}_\lambda$&$\lambda=10^3$   &$61.37_{\pm 0.68}$   &$ 74.72_{\pm 0.56}$   &$63.12_{\pm 0.72}$   &$76.35_{\pm 0.58} $ &$71.58_{\pm 0.73}$   &$85.31_{\pm  0.48}$   \\
			&$\lambda=10^4$   &$61.66_{\pm 0.68} $   &$77.80_{\pm 0.52}$    &$62.90_{\pm 0.74}$   &$78.07_{\pm 0.55}$  &$ 71.63_{\pm 0.73}$   &$\bm{86.23_{\pm 0.48}}$   \\
			&$\lambda=10^5$              &$60.64_{\pm 0.65}$   &$\bm{78.80_{\pm 0.48}}$    &$61.95_{\pm 0.71} $  &$\bm{79.50_{\pm 0.53}} $  &$70.62_{\pm 0.73}$  &$85.61_{\pm 0.48}$  \\
			\hline
			&$\lambda=10^0$    &$61.78_{\pm 0.66}$  &$74.69_{\pm 0.55}$   &$63.33_{\pm 0.71}$   &$75.79_{\pm 0.58}$ &$\bm{71.55_{\pm 0.72}}$   &$84.38_{\pm 0.49}$   \\
			&$\lambda=10^1$   &$\bm{62.04_{\pm 0.69}} $   &$74.74_{\pm 0.55}$   &$63.33_{ \pm 0.71} $   &$75.90_{\pm 0.59} $ &$70.87_{\pm 0.74}$   &$ 83.90_{\pm 0.52}$  \\
			&$\lambda=10^2$   &$61.42_{\pm 0.68} $   &$74.37_{\pm 0.54}$    &$63.51_{\pm 0.72} $   &$ 75.94_{\pm 0.59} $ &$71.36_{\pm 0.72}$   &$84.14 _{\pm 0.51}$   \\	
			$d^{(s_2)}_\lambda$&$\lambda=10^3$   &$62.12_{\pm 0.68}$   &$ 75.40_{\pm 0.55}$   &$\bm{63.45_{\pm 0.71}}$   &$76.66_{\pm 0.58}$ &$71.35_{\pm 0.72}$   &$83.92_{\pm 0.50}$   \\
			&$\lambda=10^4$   &$60.85_{\pm 0.67}$   &$77.51_{\pm 0.51}$    &$62.74_{\pm 0.72}$ &$78.34_{\pm 0.56}$  &$70.59_{\pm 0.72}$   &$\bm{85.64_{\pm 0.47}}$   \\
			&$\lambda=10^5$   &$60.42_{\pm 0.67}$   &$\bm{78.04_{\pm 0.50}}$    &$62.13_{\pm 0.72}$  &$\bm{79.02_{\pm 0.55}} $  &$69.69_{\pm 0.72}$  &$85.23_{\pm 0.48}$  \\
			
			\bottomrule[1pt]
		\end{tabular}\label{tab:As}
\end{table}

\begin{table}[h]
	\centering
	\scriptsize
		\caption{Accuracy ($\%$) comparison to exemplar model and prototype models using ResNet-12 for 5-way tasks on $\textit{mini}$ImageNet,$\textit{tiered}$-ImageNet and CIFAR-FS with 95$\%$ confidence intervals. }
		\begin{tabular}{cccccccc}
			\toprule[1pt]
			\textbf{} 
			\multirow{2}{*}{\bf{Method}} 
			&\multicolumn{2}{c}{\bf{\textit{mini}ImageNet}} 
			&\multicolumn{2}{c}{\bf{\textit{tiered}-ImageNet}} 
			&\multicolumn{2}{c}{\bf{CIFAR-FS}}\\
			
			\cmidrule(r){2-3} \cmidrule(r){4-5} \cmidrule(r){6-7}
			
			& \bf{1-shot} &\bf{5-shot} &\bf{1-shot} &\bf{5-shot} &\bf{1-shot}  &\bf{5-shot}\\
			\hline
			Exemplar ($d^{(s_1)}_\lambda$, $\lambda=0$)         &$61.21_{\pm 0.60}$ &$74.90_{\pm 0.54}$    &$62.31_{ \pm 0.72} $  &$75.07_{\pm 0.59} $  &$71.63_{\pm 0.72}$   &$83.79_{\pm 0.50}$  \\
			Prototype ($d^{(s_1)}_\lambda$, $\lambda \to\infty$)   &$60.95_{\pm 0.67} $ &$\bm{79.22_ {\pm 0.49}} $ &$60.45_{\pm 0.71}$  &$ 78.44_{\pm 0.56}$  &$70.86_{\pm 0.72}$  &$85.09_{\pm 0.50}$  \\
			SENet  ($d^{(s_1)}_\lambda, 0 < \lambda < \infty$)  &$\bm{61.81_ {\pm 0.68}} $ &$78.80_{\pm 0.48}$ &$\bm{63.27_{\pm 0.72}}$  &$ \bm{79.50_{\pm 0.53}}$  &$\bm{72.06_{\pm 0.72}}$  &$\bm{86.23_{\pm 0.48}}$  \\
			\hline
			Exemplar ($d^{(s_2)}_\lambda$,$\lambda=0$)     &$62.01_{\pm 0.68} $   &$74.42_{\pm 0.55} $    &$\bm{63.77_{\pm 0.72}}$  &$76.08_{\pm 0.57}$  &$70.93_{\pm 0.73}$   &$83.90_{\pm0.52} $  \\
				Prototype ($d^{(s_2)}_\lambda$,$\lambda \to\infty$)  &$60.85_{\pm 0.67} $ &$77.98_{\pm 0.51}$ &$62.31_{\pm 0.73} $  &$79.00 
				_{\pm 0.56}$  &$\bm{71.79_{\pm 0.72}}$  &$85.55_{\pm 0.48}$ \\
				SENet ($d^{(s_2)}_\lambda$, $0 < \lambda < \infty$)  &$\bm{62.04_ {\pm 0.69}}$ &$\bm{78.04_{\pm 0.50}}$ &$63.45_{\pm 0.27}$  &$\bm{79.02_{\pm 0.55}} $  &$71.55_{\pm 0.72}$  &$\bm{85.64_{\pm 0.47}}$ \\
				\bottomrule[1pt]
			\end{tabular}\label{tab:12}
	\end{table}

A further comparison of prototype model, exemlar model and the SENet is conducted, to illustrate the effectiveness of the proposed SENet with an appropriate $\lambda$. On $\textit{mini}$ImageNet and $\textit{tiered}$-ImageNet, the number of episodes per batch is 4 and on CIFAR-FS, the number of episodes per batch is 8. \cref{tab:12} shows the SENet loss objective with $d^{(s_1)}_{\lambda}$ or $d^{(s_2)}_{\lambda}$, compared with their corresponding Exemplar model ($\lambda = 0$) and Prototype model ($\lambda \to \infty$). It is clear that for most cases, the best performance of SENet can be obtained by setting an optimal $\lambda$ in the range of (0, $\infty$). In particular, the accuracy achieved by SENet is 1.1$\%$ higher than the prototype for 5-shot task on CIFAR-FS. These results illustrate the effectiveness of $\lambda$ setting. In addition, we found that the improvements of $d^{(s_1)}_{\lambda}$ is more obvious than those of $d^{(s_2)}_{\lambda}$ in these cases, which  support the reason for filtering the queries.

\begin{figure}[h]
	\centering
\begin{subfigure}{0.45\textwidth}
	\includegraphics[width=\linewidth]{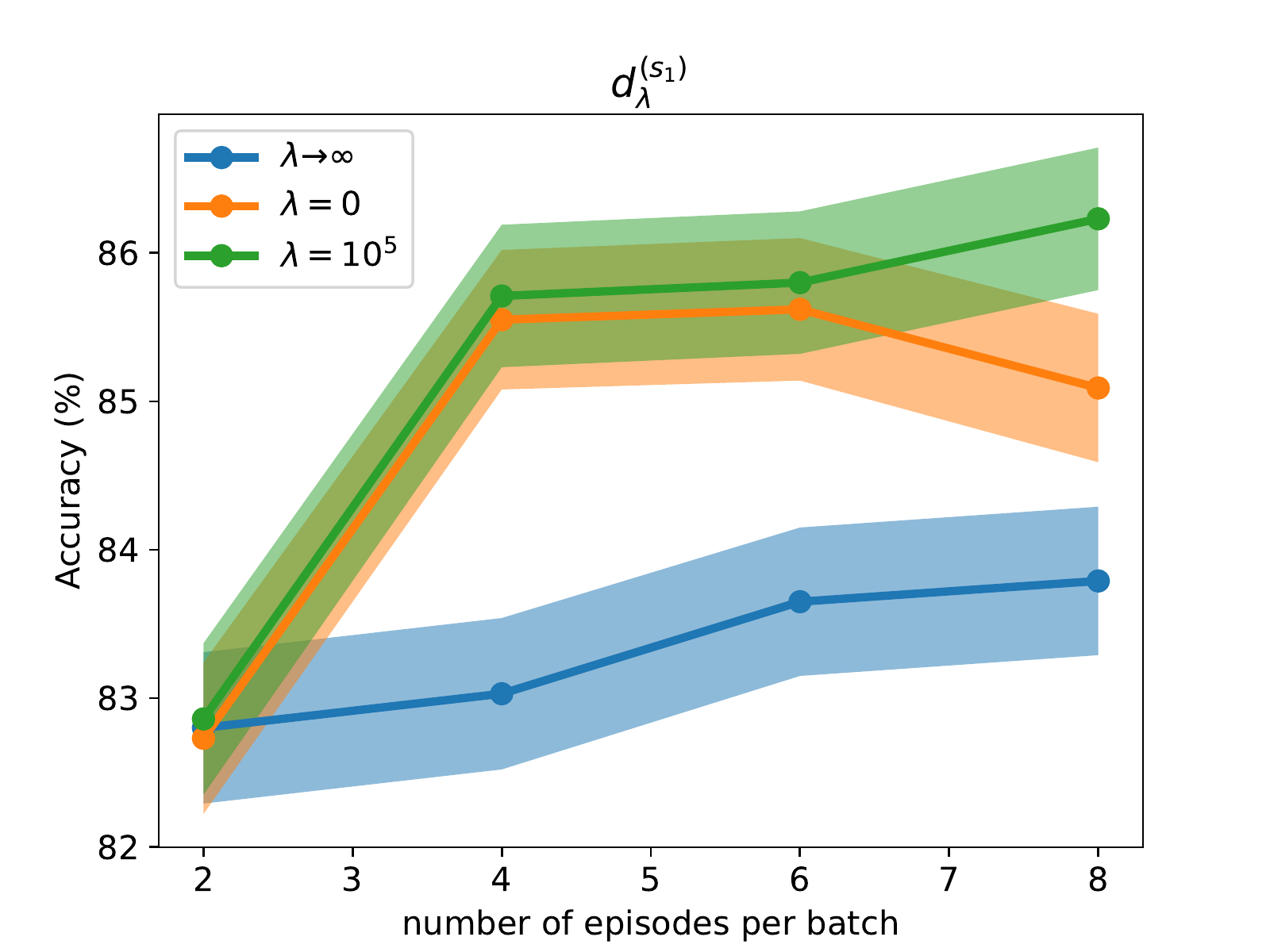}
	\caption{}
\end{subfigure}
\begin{subfigure}{0.45\textwidth}
	\includegraphics[width=\linewidth]{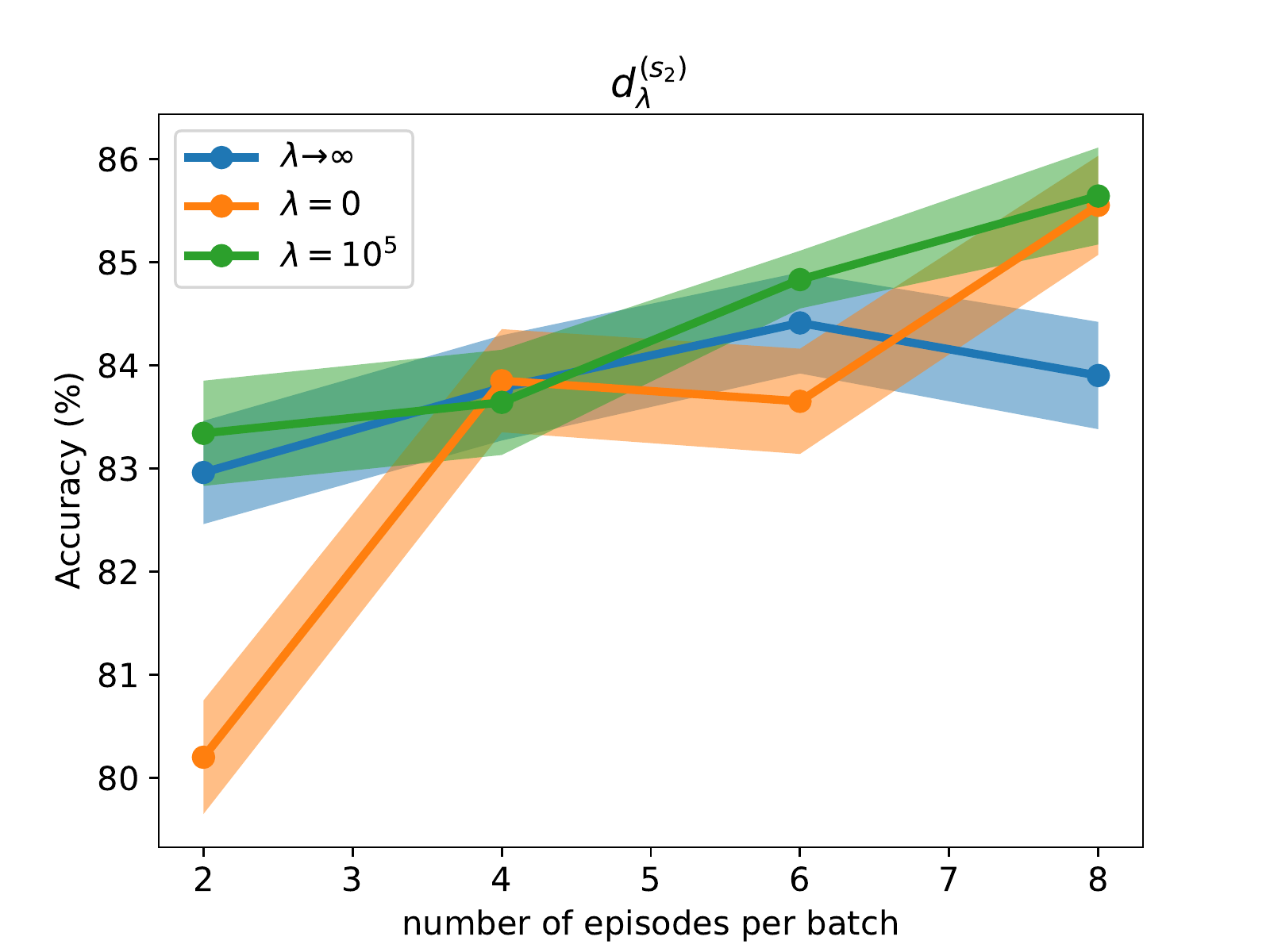}
	\caption{}
\end{subfigure}
\begin{subfigure}{0.45\textwidth}
	\includegraphics[width=\linewidth]{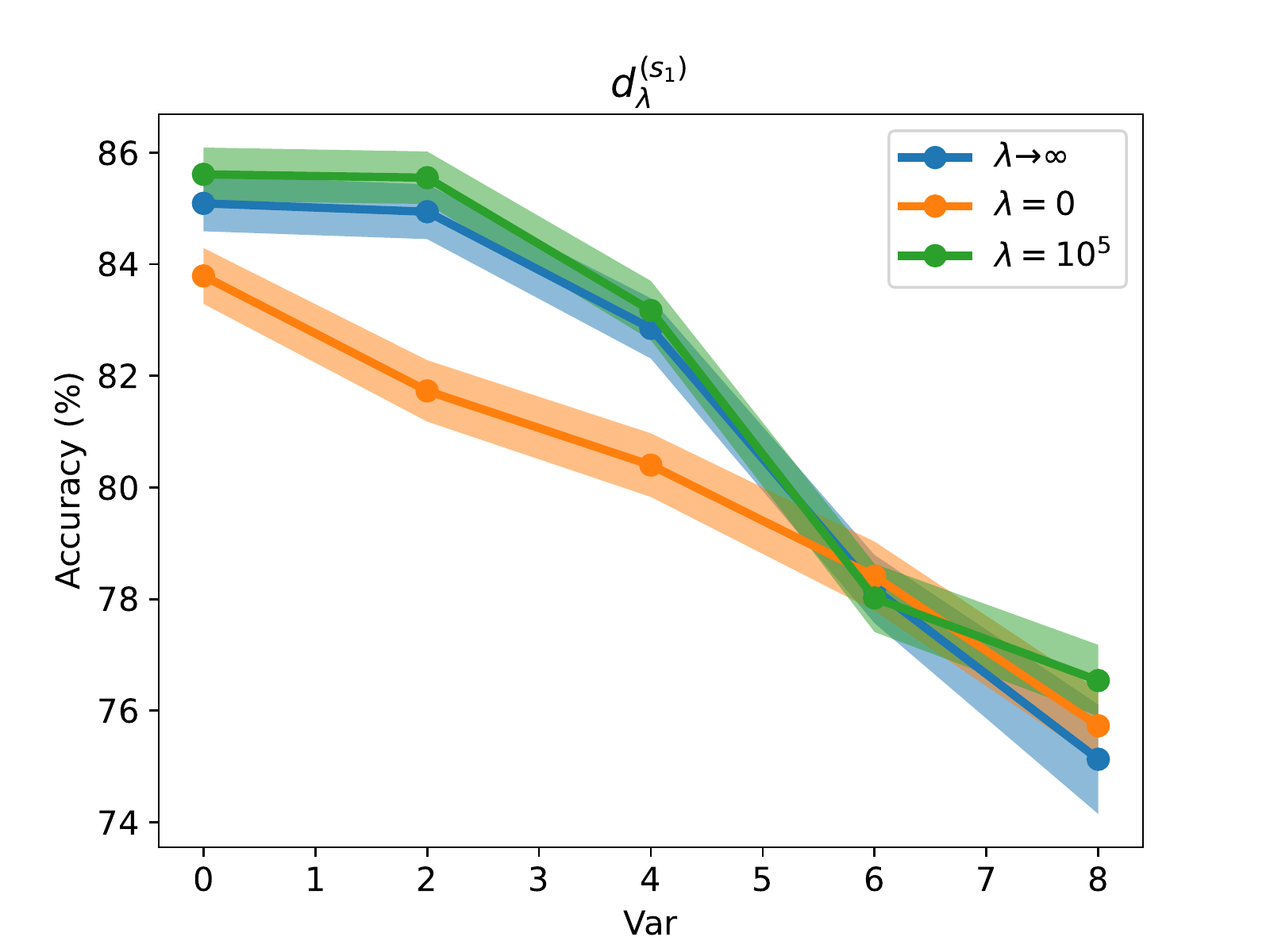}
	\caption{}
\end{subfigure}
\begin{subfigure}{0.45\textwidth}
	\includegraphics[width=\linewidth]{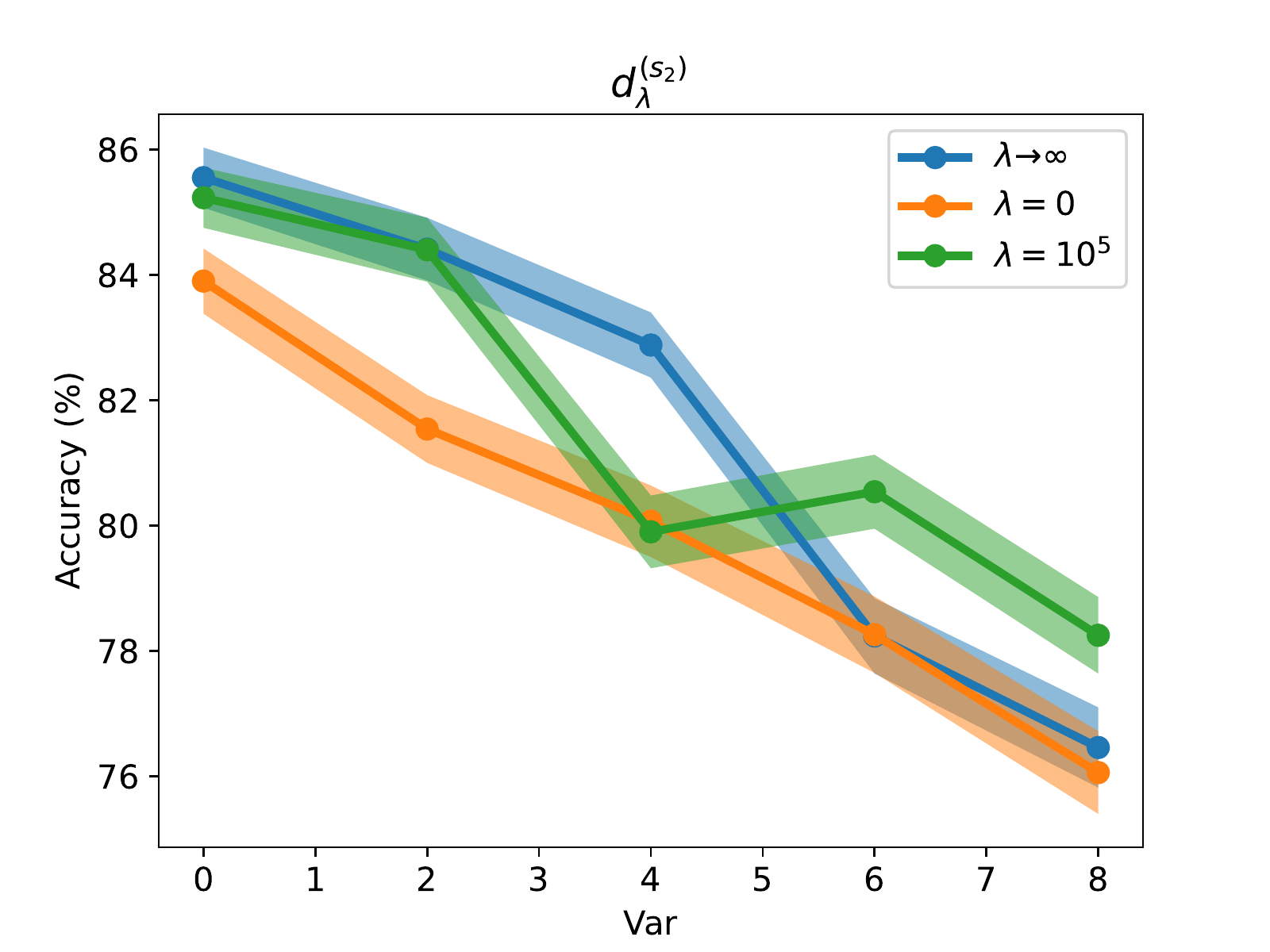}
	\caption{}
\end{subfigure}
	\caption[]{The robustness comparison of the prototype model ($\lambda \to \infty$), the exemplar model ($\lambda=0$) and the proposed SENet ($\lambda=10^5$)  on CIFAR-FS dataset. We shows accuracies achieved by (a) $d^{(s_1)}_{\lambda}$ and (b) $d^{(s_2)}_{\lambda}$ compered with the other two models respectively with different episodes per batch, and accuracies achieved by (c) $d^{(s_1)}_{\lambda}$ and (d) $d^{(s_2)}_{\lambda}$ compered respectively with the other two models on Gaussian noise with different variances (the mean is zero) on 8 episodes per batch.}
	\label{fig:Rob}
\end{figure}

\subsection{Robustness Comparison}
				
In this section, the robustness of our proposed model ($\lambda=10^5$) for few-shot learning is studied. We first compared our models with the prototype model and the exemplar model using different episodes per batch, as shown in \cref{fig:Rob}(a) and \cref{fig:Rob}(b).  We can see that our proposed model almost perform the best with all different cases of episodes per batch. Furthermore,  We compare them under Gaussian noise with different variances to show their ability of anti-noise, as shown in \cref{fig:Rob}(c) and \cref{fig:Rob}(d). We can see that as the variance is large, e. g., 8 variance,  the proposed SENet with two filters respectively shows better ability of anti-noise than the other two models.

\subsection{Highway and Large-shot Performance}
\subsubsection{Large-shot performance.} Here we compare SENet with prototype model and exemplar model with 1 episodes per batch for the large-shot tasks, e. g., 10-shot and 20-shot tasks, as listed in  \cref{tab:large}. For the SENet, we set $\lambda$ as a optimal value. It can be seen that for these tasks, the SENet can perform the best compared with other two models. A significant improvement is that, for the 20-shot task on $\textit{mini}$ImageNet, SENet achieves the accuracy of 81.30$\pm$0.42$\%$, which is about 4$\%$ higher than the other two models. Comparing these two methods, it can be seen that when selecting the appropriate $\lambda$, the overall accuracy of method $d^{(s_2)}_\lambda$ is slightly higher than that of method $d^{(s_1)}_\lambda$. These results show that as the shot is large in few-shot learning, the setting of $\lambda$ has significant impact on the performances of the two filtering methods.

\begin{table}[h]
	\centering
		\caption{Accuracy ($\%$) comparison to exemplar model and prototype models using ResNet-12 for 5-way tasks on $\textit{mini}$ImageNet and $\textit{tiered}$-ImageNet for large-shot tasks with 95$\%$ confidence intervals. }
		\begin{tabular}{cccccccc}
			\toprule[1pt]
			\textbf{} 
			\multirow{2}{*}{\bf{Method}} 
			&\multicolumn{2}{c}{\bf{\textit{mini}ImageNet}} 
			&\multicolumn{2}{c}{\bf{\textit{tiered}-ImageNet}}\\
			
			\cmidrule(r){2-3} \cmidrule(r){4-5}
			
			& \bf{10-shot} &\bf{20-shot} &\bf{10-shot} &\bf{20-shot} \\
			\hline
			Exemplar ($d^{(s_1)}_\lambda$, $\lambda=0$)         &$76.04_{\pm 0.47}$ &$77.78_{\pm 0.45}$    &$72.28_{\pm 0.55}$  &$75.46_{\pm 0.52}$  \\
			Prototype ($d^{(s_1)}_\lambda$, $\lambda \to\infty$)   &$73.43_{\pm 0.55}$ &$77.26_{\pm 0.49}$ &$71.68_{\pm 0.62}$  &$75.42_{\pm 0.55}$    \\
			SENet  ($d^{(s_1)}_\lambda, 0 < \lambda < \infty$)  &$\bm{79.02_{\pm 0.44}}$ &$\bm{81.30_{\pm 0.42}}$ &$\bm{72.54_{\pm 0.62}}$  &$\bm{76.69_{\pm 0.54}}$   \\
			\hline
			Exemplar ($d^{(s_2)}_\lambda$,$\lambda=0$)     &$74.03_{\pm 0.54}$  &$77.52_{\pm 0.49}$    &$72.74_{\pm 1.23}$ &$75.52_{\pm 1.21}$ \\
			Prototype ($d^{(s_2)}_\lambda$,$\lambda \to\infty$)  &$76.32_{\pm 1.14}$ &$78.42_{\pm 1.15}$ &$72.79_{\pm 0.60}$  &$76.75_{\pm 0.54}$   \\
			SENet ($d^{(s_2)}_\lambda$, $0 < \lambda < \infty$)   &\bm{$79.74_{\pm 0.50}}$&$\bm{80.98_{\pm 1.11}}$&$\bm{73.06_{\pm 0.55}}$  &\bm{$76.75_{\pm 0.55}}$   \\
			\bottomrule[1pt]
		\end{tabular}\label{tab:large}
\end{table}
				
\subsubsection{Highway performance.} 
\begin{figure}[h]
	\centering

\begin{subfigure}{0.9\textwidth}
	\includegraphics[width=\linewidth]{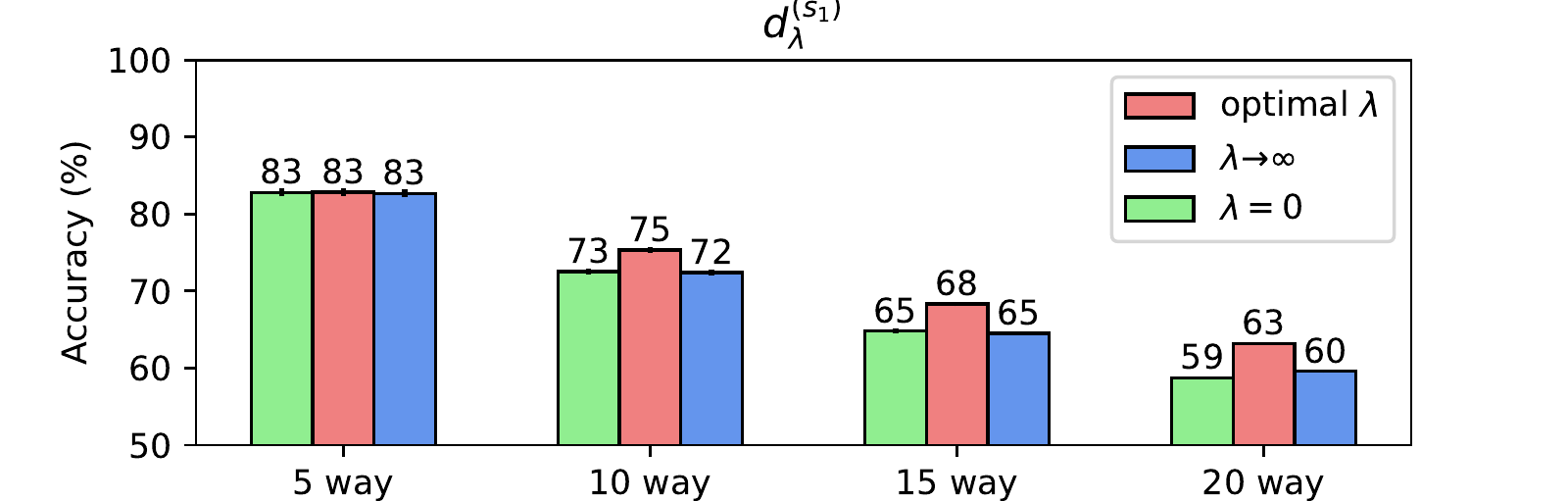}
\end{subfigure}
\begin{subfigure}{0.9\textwidth}
	\includegraphics[width=\linewidth]{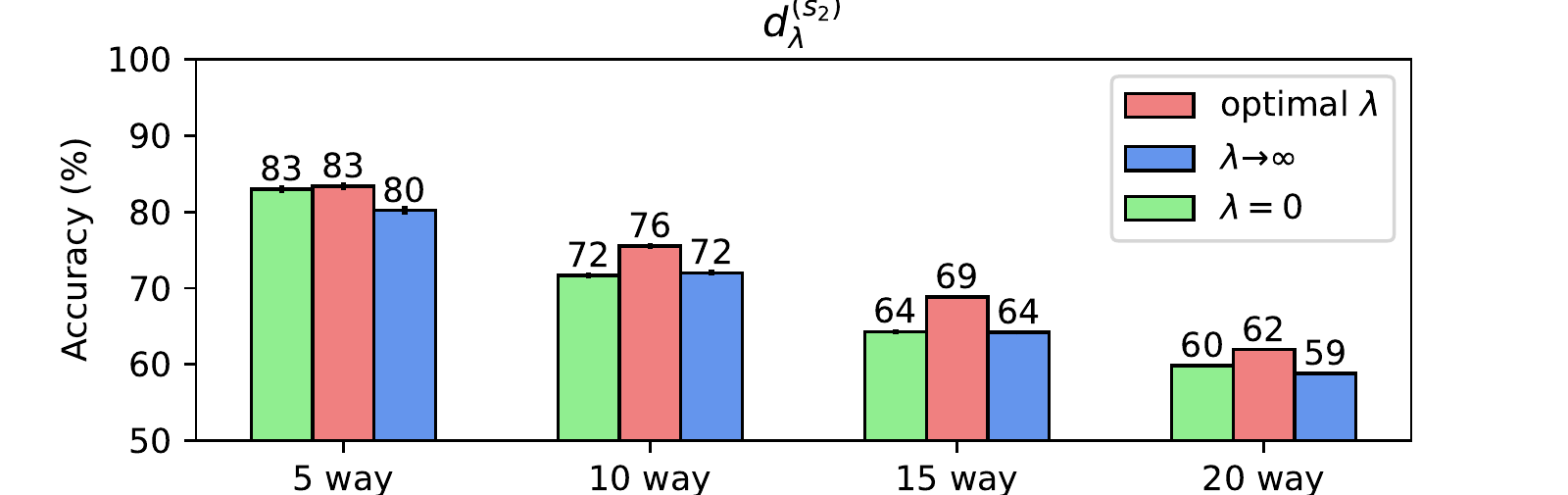}
\end{subfigure}
	\caption[]{The comparison of performance of the prototype model ($\lambda \to \infty$), the exemplar model ($\lambda=0$) and the proposed SENet with (a) $\lambda = 1000$ for $d^{(s_1)}_\lambda$ and (b) $\lambda = 10000$ for $d^{(s_2)}_\lambda$ with 2 episodes per batch on CIFAR-FS dataset.}
	\label{fig:high}
\end{figure}

Further, we compare the performance of the prototype model, the exemplar model and our proposed SENet in the case of 5-way, 10-way, 15-way and 20-way with 2 episodes per batch, in order to observe whether our model has better performance in the highway cases, as shown in \cref{fig:high}(a) and \cref{fig:high}(b). \cref{fig:high} (a) shows that the $d^{(s_1)}_\lambda$ with an optimal $\lambda$ performs the best while the way is high in despite of no improvement while the way is low. Nonetheless, it should be noted that the improvement exists for 5-way task using 8 episodes per batch (See 4.3 section). \cref{fig:high} (b) shows that the $d^{(s_2)}_\lambda$ with a optimal $\lambda$ performs better than those of exemplar model ($\lambda=0$) and those of prototype model ($\lambda \to \infty$). In particular, the accuracy of our proposed model achieves about 4.5$\%$ higher than those of other two models. these results show that the two proposed filtering strategies is effective and can perform excellent performance in the highway case.

\section{Conclusions and Future Work}
Prototype is a mainstream representation of category for few-shot learning, but lacks of the ability to flexibly adjust inductive bias. We proposed a novel model, the Shrinkage Exemplar Networks (SENet), to address this drawback of prototype utilizing the advantage of exemplar model, inspired by a debate in cognitive psychology of using whether prototype model or exemplar model for classification of things. SENet represents category with embedding of support samples that shrink to their mean via spectral filtering by using a shrinkage estimator. Two types of filters were designed, one for moderate filtering of irrelevant components (noise) and the other for appropriate enhancement of common representation of categories.We conduct several experiments to demonstrate the effectiveness of SENet for few-shot learning.
				
The ability to improve the performance of SENet via shrinkage coefficient lies in setting an appropriate value, and the model would even be sensitive to the shrinkage coefficient \cite{muandet2016kernel}. Since there may exist a room for the further improvement while adjusting shrinkage coefficient with a more exquisite way, the next research direction is to explore sophisticated empirical settings or learning methods. The design of the shrinkage estimator is also a meaningful topic \cite{muandet2016kernel}. In addition, since we have included all sample details in the model training, its time complexity is higher than simply using prototypes. Therefore, how to quickly train for the proposed loss function is also a critical problem that needs to be addressed in future work.
				
\bibliographystyle{splncs04}
\bibliography{main}
				
\end{document}